\documentclass[10pt,twocolumn,letterpaper]{article}

\usepackage{cvpr}      

\usepackage{graphicx}
\usepackage{amsmath}
\usepackage{amssymb}
\usepackage{booktabs}
\usepackage{times}
\usepackage{latexsym}
\usepackage{tabularx}
\usepackage{booktabs}
\usepackage{multirow}
\usepackage{adjustbox}
\usepackage{caption}
\usepackage{subcaption}
\usepackage{tipa} 
\usepackage{enumitem}
\usepackage{amssymb}
\usepackage{amsthm}
\usepackage{pifont} 
\usepackage{array}
\usepackage{comment}
\usepackage{amsmath}
\usepackage{mathtools}
\usepackage{algorithm}
\usepackage{algpseudocode}

\usepackage[pagebackref,breaklinks,colorlinks]{hyperref}

\usepackage[capitalize]{cleveref}
\crefname{section}{Sec.}{Secs.}
\Crefname{section}{Section}{Sections}
\Crefname{table}{Table}{Tables}
\crefname{table}{Tab.}{Tabs.}

\def\cvprPaperID{*****} 
\def\confName{CVPR}
\def\confYear{2022}

\begin{document}

\title{A Study of Acquisition Functions for Medical Imaging Deep Active Learning}

\author{\normalsize Bonaventure F. P. Dossou\\ 
{\tt\small bonaventure.dossou@mila.quebec}\\
\footnotesize
Mila Quebec AI Institute, McGill University, Masakhane NLP, Lelapa AI}

\maketitle

\begin{abstract}
   The Deep Learning revolution has enabled groundbreaking achievements in recent years. From breast cancer detection to protein folding, deep learning algorithms have been at the core of very important advancements. However, these modern advancements are becoming more and more data-hungry, especially on labeled data whose availability is scarce: this is even more prevalent in the medical context. In this work, we show how active learning could be very effective in data scarcity situations, where obtaining labeled data (or annotation budget is very limited). We compare several selection criteria (BALD, MeanSTD, and MaxEntropy) on the ISIC 2016 dataset. We also explored the effect of acquired pool size on the model's performance. Our results suggest that uncertainty is useful to the Melanoma detection task, and confirms the hypotheses of the author of the paper of interest, that \textit{bald} performs on average better than other acquisition functions. Our extended analyses however revealed that all acquisition functions perform badly on the positive (cancerous) samples, suggesting exploitation of class unbalance, which could be crucial in real-world settings. We finish by suggesting future work directions that would be useful to improve this current work. The code of our implementation is open-sourced at \url{https://github.com/bonaventuredossou/ece526_course_project}   
\end{abstract}

\section{Introduction}
\label{sec:intro}
Active learning (AL) is generally defined as a semi-supervised machine learning (ML) algorithm whose goal is to use relatively few initial training samples in order to achieve better performance of a given model $\mathcal{M}$. The optimization of $\mathcal{M}$ is done by iteratively training it and making it learn how to choose useful new data samples to label, from a pool of unlabelled data, which will help it find better parameters and improve its overall performance on downstream tasks (e.g., prediction accuracy). The query and acquisition of new samples from the pool of unlabeled data are often done using uncertainty-based measures \cite{bayesiancnn}, and selecting the most uncertain samples in the pool of unlabeled data samples. Due to the fact that AL-based methods learn to \textit{smartly} pick useful samples for their learning, this makes AL a prevalent paradigm to cope with data scarcity (which is often a bottleneck to many ML applications (e.g. in the medical where patient data is rare, sensitive, and subject to many privacy issues). The efficiency of active learning (i.e. its ability to produce better performance despite being trained on smaller training data) has been proven in many works of literature.

In \cite{jain2022biological}, the authors have explored active learning for biological sequence design. The empirical results have proven that active learning coupled with uncertainty-based acquisition functions (Upper Confidence Bounds (UCB) \cite{Srinivas2009GaussianPO}, Expected Improvement (EI) \cite{Mockus1974OnBM}) enables the discovery, and generation of novel and diverse biological compounds (e.g. antimicrobial peptides (AMPs), green fluorescent proteins, and DNA sequences with high binding signals). This has also been confirmed in \cite{Dossou2022GraphBasedAM} using active learning and graph-connected components. In \cite{al_ner, al_ner_2, ein-dor-etal-2020-active, siddhant2018deep}, for the task of clinical named entity recognition (C-NER), the authors have shown that with as little as 50\% of the initial training data, active learning still achieve high performance ($\sim$ 99\% of the accuracy of tokens predictions). Finally, in language modeling \cite{dossou-etal-2022-afrolm}, built an active learning-based language model (from scratch) for 23 African languages. The resulting language model called AfroLM (trained on less than 1GB of training data) outperformed existing state-of-the-art models like BERT \cite{Devlin2019BERTPO}, XLMR \cite{Conneau2019UnsupervisedCR}, and AfriBERTa \cite{Ogueji2021SmallDN} (all trained on terabytes of data) on NER, Text Classification, and Sentiment Analysis tasks.

In this work, we are exploring epistemic uncertainty (hereafter referred to as \textit{uncertainty}), which refers to the uncertainty of the model in low-resource (lack of training data or availability of a very small amount of data) settings. In order to get some uncertainty score, most existing works make use of kernel-based methods on pair of images in order to capture image similarity \cite{Zhu2003SemiSupervisedLU, Li2013AdaptiveAL, Joshi2009MulticlassAL}. Conversely to these methods, as in the paper we trying to explore in this paper, we will make use of Bayesian CNNs \cite{Gal2015BayesianCN} which are ``\textit{Convolutional Neural Networks (CNNs) \cite{LeCun1989BackpropagationAT} with prior probability distributions placed over a set of model parameters}``\cite{bald}. In the original paper titled ``\textit{Deep Bayesian Active Learning with Image Data}``\cite{bald}, authors demonstrated the use and advantage of Active Learning based methods, using the MNIST dataset \cite{LeCun2005TheMD}. Moreover, the authors have shown that Active Learning based methods using uncertainty criteria converged faster and performed better than Active Learning using other types of selection criteria (e.g. MBR \cite{Zhu2003SemiSupervisedLU} which uses an RBF kernel over the raw images to get a similarity graph which can be used to share information about the unlabelled pool).

Given the success of their method, the authors stated ``\textit{propagating uncertainty throughout the model, helps the model attain higher accuracy early on and converge to a higher accuracy overall. This demonstrates that the uncertainty propagated throughout the Bayesian models has a significant effect on the models’ measure of their confidence.}`` Additionally, as the authors explored different acquisition functions, they stated that \textbf{\textit{BALD}} will encourage the selection (by the model) of set points (we can think of this as an augmentation set, a set with unlabeled data) that are expected to maximize the information gained about the model parameters (or in another word, maximize the mutual information between predictions and model posterior).

In this report, exploring the ISIC 2016 Melanoma Diagnosis dataset \cite{Gutman2016SkinLA} we attempted to answer the following questions:
\begin{itemize}
    \item is model uncertainty really beneficial to the Melanoma detection task (disguised here as a binary classification)?
    \item is it more efficient for medical imaging in general, and in particular for the Melanoma Detection task, to query and acquire the most uncertain samples or least uncertain samples?
    \item which acquisition function or selection criteria works better for the Melanoma Detection task? Is it BALD as the authors claimed?
    \item what is the effect of the size of the set of newly acquired data points, on the model's overall performance? 
\end{itemize}
In order to answer the following questions, in the following sections we: \begin{enumerate}
    \item describe the different acquisition functions explored in the paper of interest
    \item describe the dataset and the task at hand
    \item provide implementation details (hyperparameters, etc)
    \item report the different results of analyses and ablation studies performed with a discussion about the results.
\end{enumerate}

\section{Acquisition Functions}
In this section, we describe the different acquisition functions we implemented, from the paper of interest.
\subsection{Maximum Entropy}
This acquisition function or selection criteria aims at selecting the data points which maximize the entropy of the model over each unlabelled data sample and known labels (classes). With the entropy defined as 
\[\mathcal{H}[y|x, D_{T}] = -\sum_{c}p(y=c|x, D_{T})logp(y=c|x, D_{T})\] where $D_{T}$ is the training set, which is augmented by the set of newly acquired samples at each active learning round.
\subsection{Mean Standard Deviation}
The Mean Standard Deviation (for short MeanSTD) is the most commonly used acquisition function. It leverages the variance of the model over classes, given an input $x$ and the parameters $w$ of the model\cite{Kampffmeyer2016SemanticSO, Kendall2015BayesianSM}. It is mathematically defined as follow: \[\sigma_{c} = \sqrt{\mathcal{E}_{q(w)}[p(y=c|x, w)^{2}] - \mathcal{E}_{q(w)}[p(y=c|x, w)]^{2}}\]
\[\sigma_{x} = \frac{1}{C}\sum_{c} \sigma_{c}\]
As with the Maximum Entropy, in this scheme, we are also selecting points that maximize the MeanSTD.
\subsection{BALD}
BALD \cite{Houlsby2011BayesianAL} is based on mutual information. By definition, the mutual information denoted $\mathcal{I}$ between two random variables $X, Y$ is telling us ``\textit{how much uncertainty do we observe in X if we observe Y}``. BALD focuses on maximizing the mutual information between the predictions of the model and its posterior. BALD is mathematically defined as
\[\mathcal{I}(y, w|x, D_{T}) = \mathcal{H}[y|x, D_{T}] - \mathcal{E}_{p(w|D_{T})}[\mathcal{H}[y|x, w]]\]
\[ = -\sum_{c}p(y=c|x, D_{T})logp(y=c|x, D_{T})\]
\[+ \mathcal{E}_{p(w|D_{T})}[\sum_{c}p(y=c|x, w)logp(y=c|x, w)]\] where $w$ are the parameters of the model. In other words, as the authors stated ``\textit{BALD chooses points that are expected to maximize the information gained about the parameters of the model $w$} \cite{bald}``. These points are points on which the model is uncertain on average, but about which some parameters produce disagreeing predictions with high certainty.
\begin{figure}[!ht]
    \includegraphics[width=0.5\textwidth]{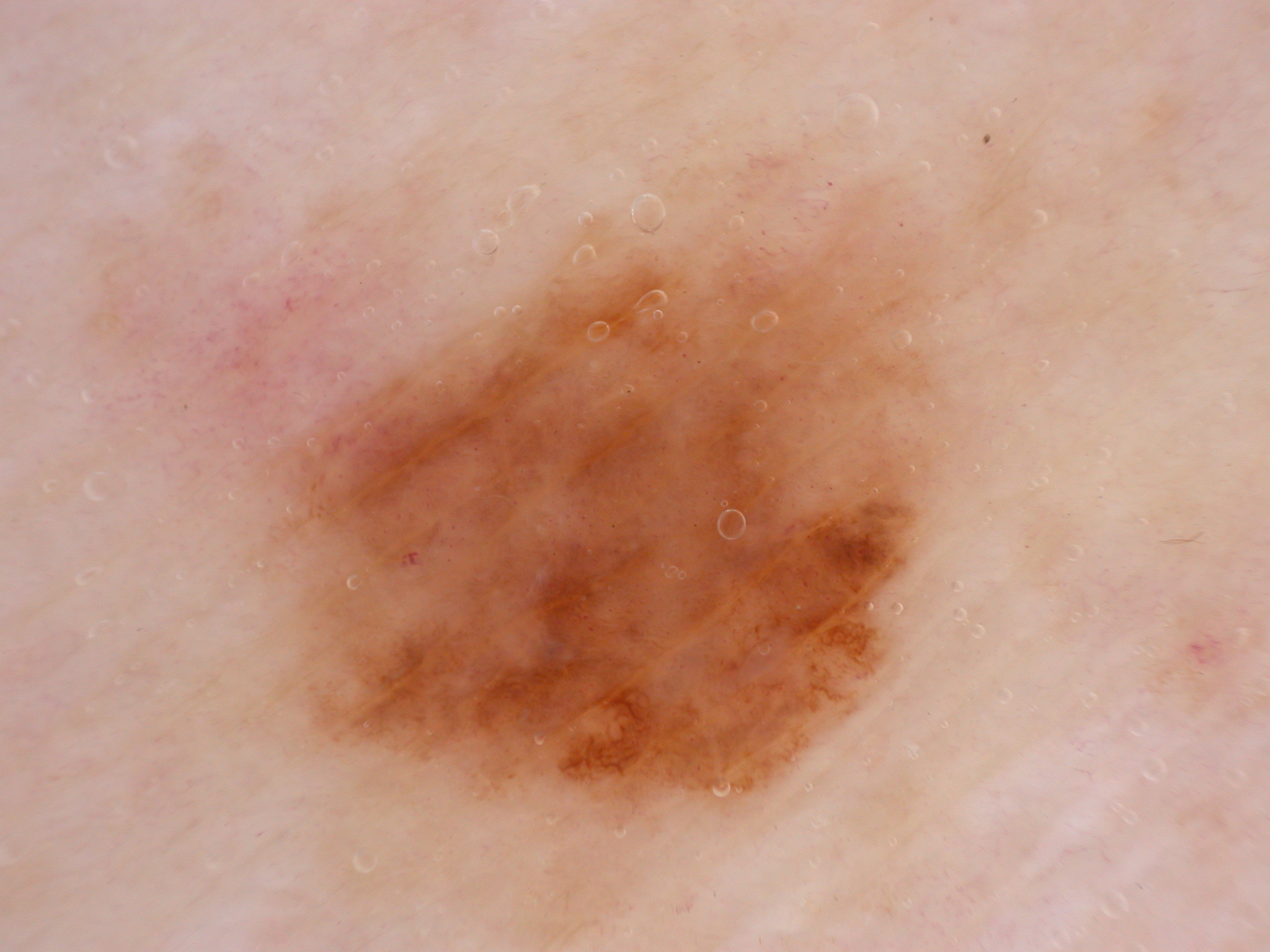}
    \caption{Picture of a non-cancerous Skin (Negative Sample)}
    \label{non_cancerous_skin}
\end{figure}
\begin{figure}[!ht]
    \includegraphics[width=0.5\textwidth]{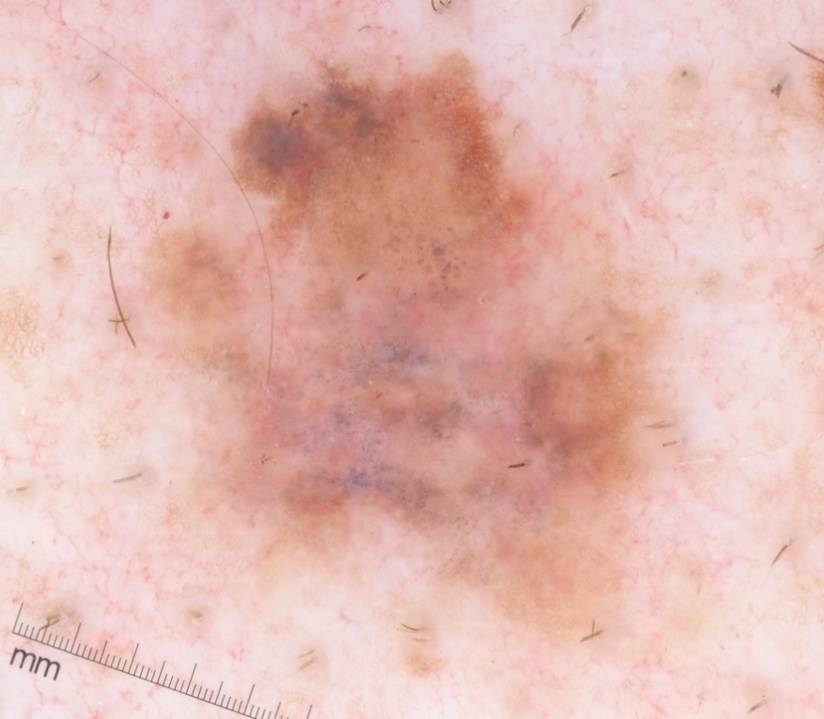}
    \caption{Picture of a cancerous Skin (Positive Sample)}
    \label{cancerous_skin}
\end{figure}
\section{Dataset and Task Description}
\begin{figure}[!ht]
    \includegraphics[width=0.5\textwidth]{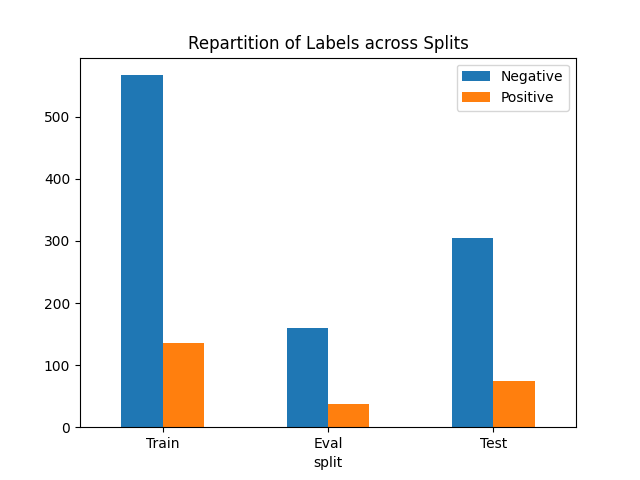}
    \caption{Repartition of Classes across Training, Validation, and Testing splits}
    \label{class_rep}
\end{figure}

The ISIC 2016 dataset \cite{Gutman2016SkinLA} has been created for the ISIC 2016 challenge. Its goal was to foster the development of image analysis tools to enable the automated diagnosis of melanoma from dermoscopic images. The ISIC 2016 dataset contains 900 training images, and 350 testing images; a rather small dataset. The task is a binary classification, with the goal to detect whether a given picture is cancerous or not (see Figures \ref{non_cancerous_skin}, and \ref{cancerous_skin}). The initial training dataset has been randomly split into two sets: training (containing 700 images) and evaluation (containing 200 images). As Figure \ref{class_rep} shows, the repartition of the classes is unbalanced, with a clear and net dominance of negative samples (non-cancerous image samples). The images are colorful (RGB format), and we have downsampled them to the shape of (224, 224).

\begin{table}[h!]
 \footnotesize
 \begin{center}
    \resizebox{0.3\textwidth}{!}{
   \begin{tabular}{cc}
   \toprule
\textbf{Hyper-parameters} & \textbf{Value}\\\hline
\# of channels & 3 \\
\# of filters & 32 \\
pooling size & 2 \\
kernel size & 4 \\
dense layer size & 128 \\
number of classes & 2 \\
dropout rate 1 & 0.25 \\
dropout rate 2 & 0.50 \\
activation function & relu \\
MC-Dropout forward passes & 20 \\
top-k & 100 \\
learning rate & 1e-4 \\
\# of epochs & 100 \\
batch size & 8 \\
image size & 224 \\
p & 0.5 \\
$l^{2}$ & 0.5 \\
optimizer & adam \\
loss function & categorical cross-entropy \\
\bottomrule
\end{tabular}
} \caption{Summary Table of Hyper-parameters}
\label{hyperpara}
\end{center}
\end{table}

\section{Hyper-parameters and Experiments}
We built the experiments using the details provided in the paper of interest \cite{bald}. Given the small size of the dataset, we started out with a small set of 100 examples made of 80 positives, and 20 negatives. Each example from all splits has been resized as stated above and normalized. The training images have additionally been augmented with \textit{Center Cropped} and \textit{Random Horizontal Flip} transformations.

The CNN architecture is made of two 2-dimensional convolutional layers, each followed by a \textit{relu} activation function. After the second \textit{convolution-activation}, the result is fed to a maximum pooling layer, followed by a dropout. The result is flattened and fed to a fully-connected layer, with later on passed successively through a dropout layer, and classification head (technically another dense layer with output dimension 2).

The network has been trained for 100 epochs, with a batch size of 8 and a learning rate of 1e-4. As the authors stated in the paper, we used Adam optimizer with weight decay \[w = \frac{(1-p)*l^{2}}{|D_{T}|}\] with $p = 0.5$ being the dropout probability, and $l^{2}$ being the length scale, set to $0.5$. At each active learning round, with a given acquisition function, we perform 20 MC-Dropout forward passes. The $top-k$, $(k=100)$ most informative samples according to the given acquisition function, are selected, added to the training set, and deleted from the pool of unlabelled data. The summary of the hyper-parameters is provided in Table \ref{hyperpara}. These hyper-parameters are kept identical across all CNN-based models for each acquisition function explored in the paper of interest and in this report.
\section{Results and Discussion}
\begin{figure}[!ht]
    \includegraphics[width=0.5\textwidth]{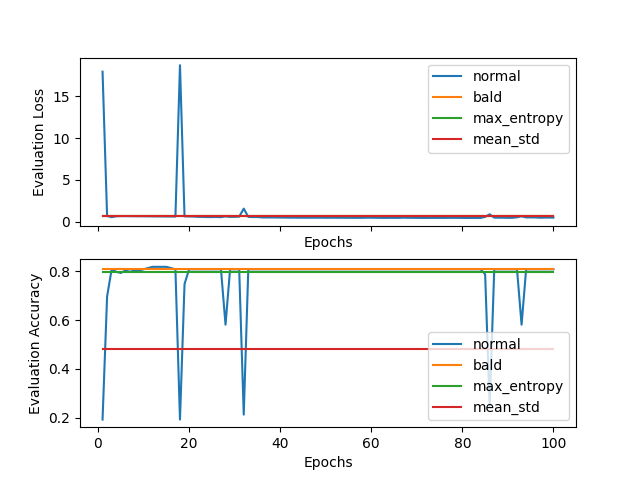}
    \caption{Evaluation Results of Bayesian CNNs with and without uncertainty}
    \label{normal_vs_final_al_rounds}
\end{figure}

Our first analysis consisted of checking the importance of uncertainty, in the context of our task description. In order to achieve that, we compared the evaluation losses and accuracies of 4 Bayesian CNNs with and without uncertainty. The results are plotted in Figure \ref{normal_vs_final_al_rounds}. The \textit{normal} legend corresponds to the model without uncertainty. From the upper subplot, we can observe that uncertainty is important to have a stable training loss, which helps, in turn, to have better performance (lower subplot). Even though for instance the accuracy performance of \textit{mean\_std} is lower than the accuracy of the normal Bayesian CNN, we can still observe given stability. These results are confirmed by the results on the testing set (see Table \ref{normal_cnn_without}). The performance of the mean\_std could intuitively make sense since the method is designed to maximize the variance of the model which could be seen as noise. This noise, coupled with the unbalanced dataset has an impact on the robustness and predictive accuracy. On the other hand, \textit{bald} maximizes the mutual information between predictions and model posterior, which over time should make the model more accurate in predicting the right class or label while being robust and coping with the data unbalance.

The surprising aspect for us came from the maximum\_entropy (interchangeably referred to as max\_entropy), which turns out to be also very efficient and stable. Theoretically, a higher entropy (since here we are maximizing it) means lower information gain: this can be considered a bit as the opposite of the goal of \textit{bald}. With the assumption that images from the same class share some specific features, we believe this behavior makes sense. This is in a way that as the model gets more exposed to non-cancerous images during training, it is more confident about them, thus learning to select samples (images) from the minority (cancerous) class. Therefore, as the active learning rounds go, the model gets more and more confident about samples from both classes and is more robust in performance.

\begin{figure}[!ht]
\includegraphics[width=0.5\textwidth]{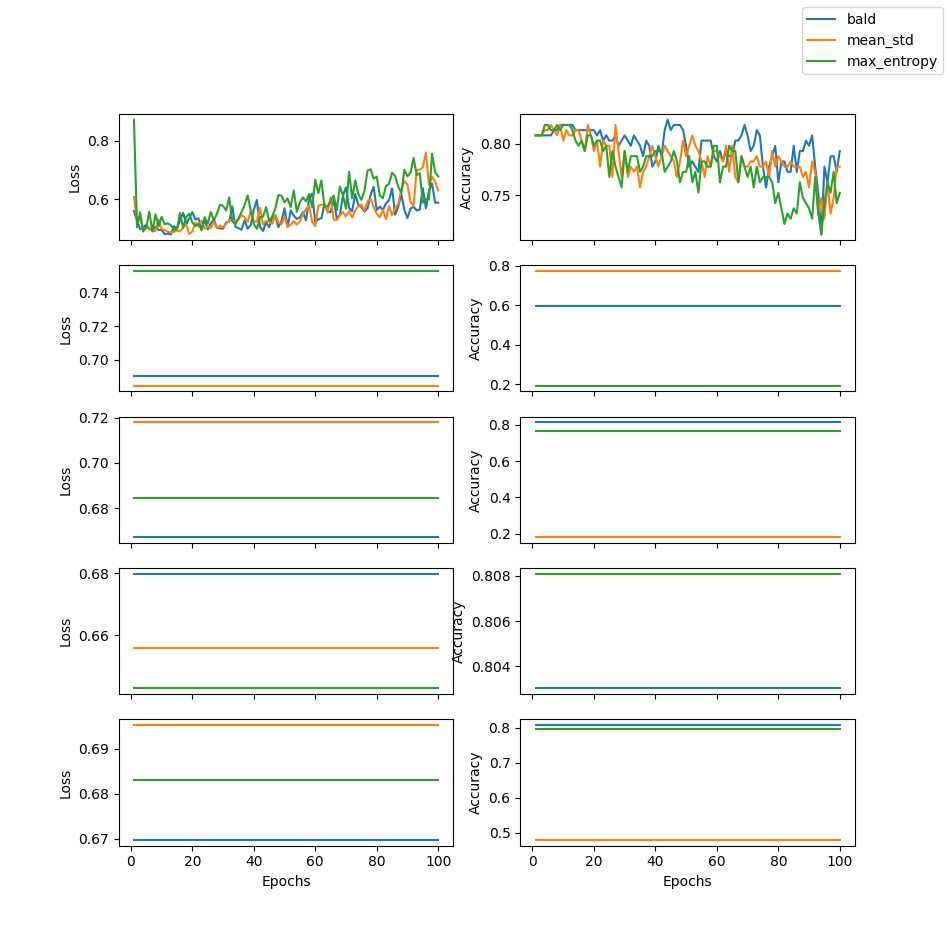}
    \caption{Evaluation Results of Bayesian CNNs across Active Learning Rounds (from top to bottom)}
    \label{al_rounds}
\end{figure}

\begin{figure}[!ht]
\includegraphics[width=0.5\textwidth]{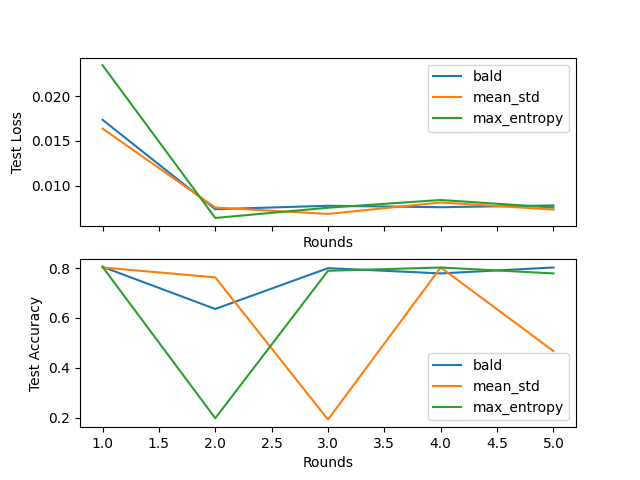}
    \caption{Test Results of Bayesian CNNs across Active Learning Rounds}
    \label{al_test_rounds}
\end{figure}

\begin{table}[h!]
 \footnotesize
 \begin{center}
    \resizebox{0.3\textwidth}{!}{
   \begin{tabular}{ccc}
   \toprule
\textbf{Method} & \textbf{Testing Loss} & \textbf{Testing Accuracy}\\\hline
normal & 0.01538 & 0.8021 \\
bald & 0.0077 & \textbf{0.8047} \\
 max\_entropy & 0.0075 & 0.7784 \\
 mean\_std & \textbf{0.0072} & 0.4670 \\
\bottomrule
\end{tabular}
} \caption{Results on the testing set for both with and without uncertainty Bayesian CNN}
\label{normal_cnn_without}
\end{center}
\end{table}
The next step of our analysis is to evaluate how different acquisition functions behave across different active learning rounds, and how the performances of the model vary; the results are in Figure \ref{al_rounds} and \ref{al_test_rounds}. As explained before, for each method, we start with 100 examples: 80 positive examples, and 20 negative examples. At each active learning round, 100 new data points are acquired (or selected according to the acquisition function) and added to the training set for the next active learning round. We performed a total of five active learning rounds. We can see that on average, once again bald performs the best. The performance of the mean\_std approach decreases over time, while the performance of the max\_entropy approach improves over the active learning rounds.

In Figure \ref{al_rounds}, we can also notice that after the first active learning round, we have more of flat lines for both the evaluation loss and accuracy. We speculate that after the first active learning round, the model underfit, and is not able to learn correctly, even though across acquisition functions we could see some variations in the value of those metrics. This could be due to the small or low capacity and complexity of the model i.e. we built very simple CNNs, with very few parameters and very few data points at each round (or even overall). Therefore there is no \textit{new drastic learning} with newly acquired data points (even though as we stated above, and as per Figure \ref{al_rounds}, we can observe some increase or decrease in performance). A way of fighting underfitting is to increase the capacity and complexity of the model. This is not something we explored in this report, but that could potentially be an extended future work.

In Figure \ref{al_test_rounds}, at first glance, we observe a very unstable performance (accuracy) of the mean\_std acquisition function, which aligns with our previous analyses of the method. The other two acquisition methods both suffer from a decreasing jump (even though the gap is bigger and more drastic for max\_entropy than for bald) before regaining high performance and remaining stable across the rounds. 

From the figure \ref{al_test_rounds}, we can still observe the better performance, and stability that \textit{bald} provides (closely approximated by maximum\_entropy), as opposed to mean\_std.

Next, we focused on evaluating whether selecting the most uncertain samples is the most beneficial way. In order to do that, we ran the same experiences but selected at each acquisition round the least uncertain samples.
\begin{figure}[!ht]
\includegraphics[width=0.5\textwidth]{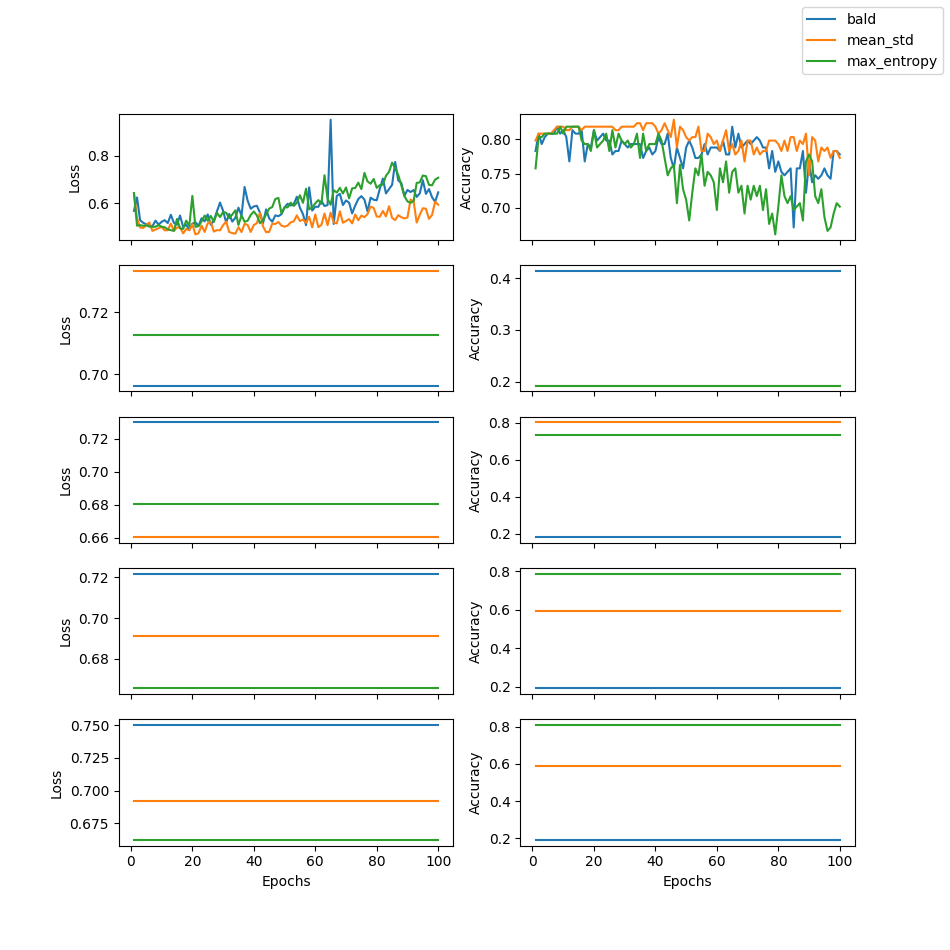}
    \caption{Evaluation Results of Bayesian CNNs across Active Learning Rounds using the Least Uncertain Samples}
    \label{al_test_rounds_least}
\end{figure}

\begin{figure}[!ht]
\includegraphics[width=0.5\textwidth]{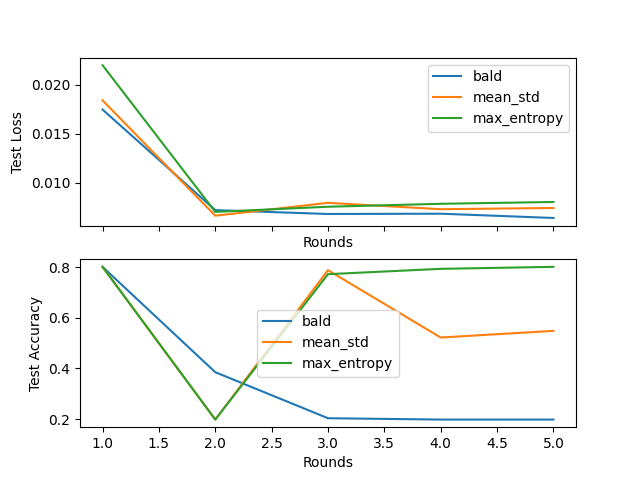}
    \caption{Test Results of Bayesian CNNs across Active Learning Rounds using the Least Uncertain Samples}
    \label{al_test_rounds_least_false}
\end{figure}

In Figure \ref{al_test_rounds_least} (as opposed to Figure \ref{al_test_rounds}), we can observe a more \textit{unstable} training, and overall a higher loss value (by the end of the five active learning rounds). On the accuracy metric, we can see that on average, \textit{bald} performed worse, which makes sense. The mean\_std has a better performance as opposed to the setting presented in Figure \ref{al_test_rounds}. This is because the points selected are the ones minimizing the variance, thus inducing less noise and encouraging better performance. The maximum\_entropy kept a relatively normal balance and suggests that it is agnostic of the sampling mode (least uncertain or most uncertain samples i.e. samples respectively minimizing or maximizing the entropy). Our results, insights, and analyses are confirmed by the results on the testing set, presented in Figure \ref{al_test_rounds_least_false}.

Thus far, our experiments, insights, and analyses have shown that:
\begin{enumerate}
    \item uncertainty is beneficial to our Melanoma Detection task
    \item in the spirit of reproducing the results of the paper of interest, we can confirm that \textit{bald} is the best acquisition function as the authors claimed
    \item our additional ablation studies have also revealed that, in the context of our Melanoma Detection task, maximum\_entropy has been proven to be agnostic of the sampling mode, offering more robustness and flexibility
\end{enumerate}
\begin{table}[!ht]
 \footnotesize
 \begin{center}
    \resizebox{0.5\textwidth}{!}{
   \begin{tabular}{ccccccccc}
   \toprule
\textbf{Method} & \textbf{Metric} & \textbf{Query=115} & \textbf{Query=100} & \textbf{Query=90} &  \textbf{Query=80}& \textbf{Query=70}& \textbf{Query=60}& \textbf{Query=50}\\\hline\hline
bald & loss & 0.0177 & 0.0174 & 0.0183 & \textbf{0.0169} & 0.0208 & 0.0201 & 0.0185  \\
bald & accuracy & 0.8047 & \textbf{0.8047} & 0.7994 & 0.7942 & 0.7994 & 0.8021 & 0.8047 \\\hline\hline
max\_entropy & loss & 0.0173 & 0.0235 & 0.0192 & 0.0202 & \textbf{0.0157} & 0.0167 & 0.0170 \\
max\_entropy & accuracy & 0.7995 & \textbf{0.8047} & 0.8021 & 0.7863 & 0.8021 & 0.8021 & 0.7863 \\\hline\hline
mean\_std & loss & 0.0185 & \textbf{0.0164} & 0.0191& 0.0177 & 0.0164 & 0.0202 & 0.0186 \\
mean\_std & accuracy & 0.7916 & 0.8021 & 0.8021 & 0.8047 & 0.7889 & \textbf{0.8074} & 0.7968 \\
\bottomrule
\end{tabular}
} \caption{Report of Testing Loss and Testing Accuracy on ISIC 2016 dataset as a function of the different query sizes. For each method and for each metric, the number in bold represents the best value achieved for a given query size.}
\label{query_size_effect}
\end{center}
\end{table}

\begin{figure}[!ht]
\includegraphics[width=0.5\textwidth]{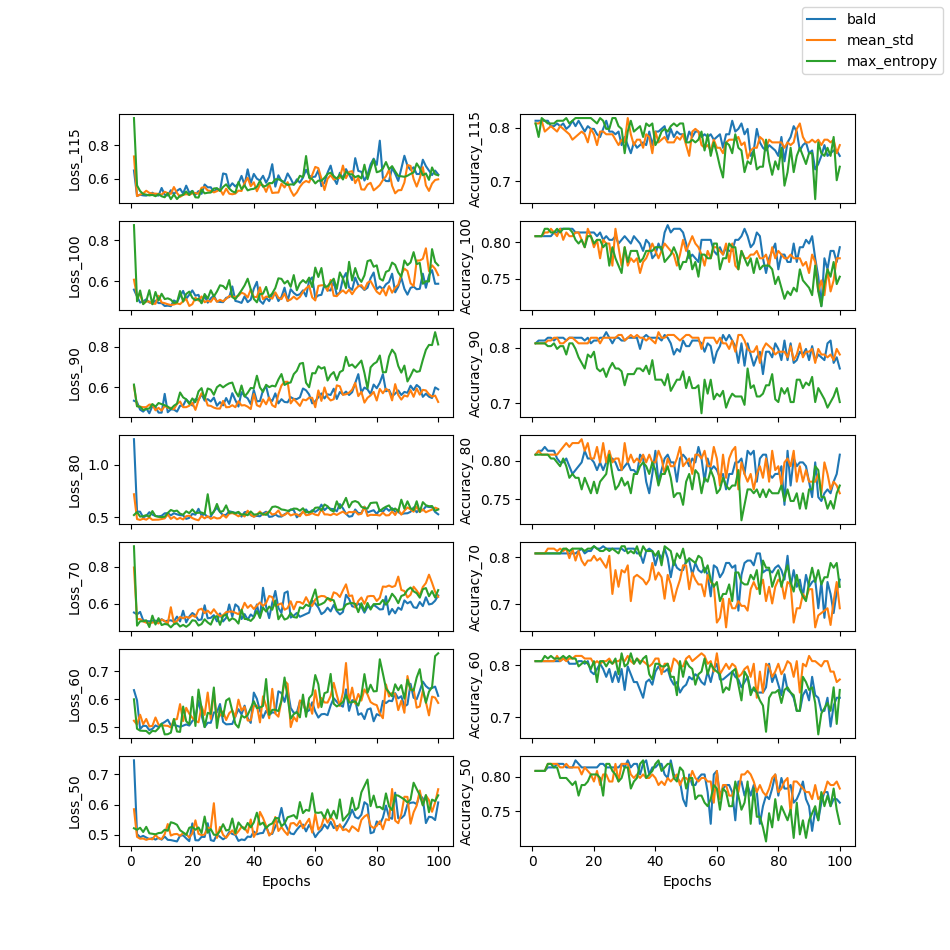}
    \caption{Impact of query size on Evaluation Loss and Accuracy. Each plot has on the y-axis a legend of the format \textit{{metric}\_{query\_size}}. The original query size is 100. Additionally, we explored different query sizes: 115, 90, 80, 70, 60, and 50.}
    \label{al_query_size}
\end{figure}

Finally, we proceeded to leverage the influence of the size of the newly acquired samples (query size). The original paper used a default of 100. In our ablation study, we tried different additional query sizes: 115, 90, 80, 70, 60, and 50. Our previous experiments revealed that the \textit{learning} happens mainly on the first active learning round. Therefore, we focused on the impact of the query sizes, solely in the first active learning round. The results are presented in Figure \ref{al_query_size} and Table \ref{query_size_effect}. 

In Figure \ref{al_query_size}, we can notice that the scale of the loss and accuracy does not change that much. However, as far as the loss metric is concerned, we can observe that generally, all acquisition functions (\textit{bald}, mean\_std, and in particular max\_entropy) are impacted by the query size. On the accuracy scale, we can see that max\_entropy brings more fluctuations or noise, as the query size decreased. This is the same (but on a lower scale) for mean\_std, while \textit{bald} has more stability. As the authors stated, this \textit{``might be because BALD avoided selecting noisy points: nearby images for which there exist multiple noisy labels of different classes. Such points have large aleatoric uncertainty – uncertainty which cannot be explained away – rather than large epistemic uncertainty – the uncertainty which BALD captures in order to explain it away, i.e. reduce it``}\cite{bald}. Moreover, we can see that the accuracy results are very similar across query sizes and acquisition methods, on the fixed test set. This demonstrates the difficulties with handling ML performances of ML models in extremely small data regimes.

\begin{figure}[!ht]
\includegraphics[width=0.5\textwidth]{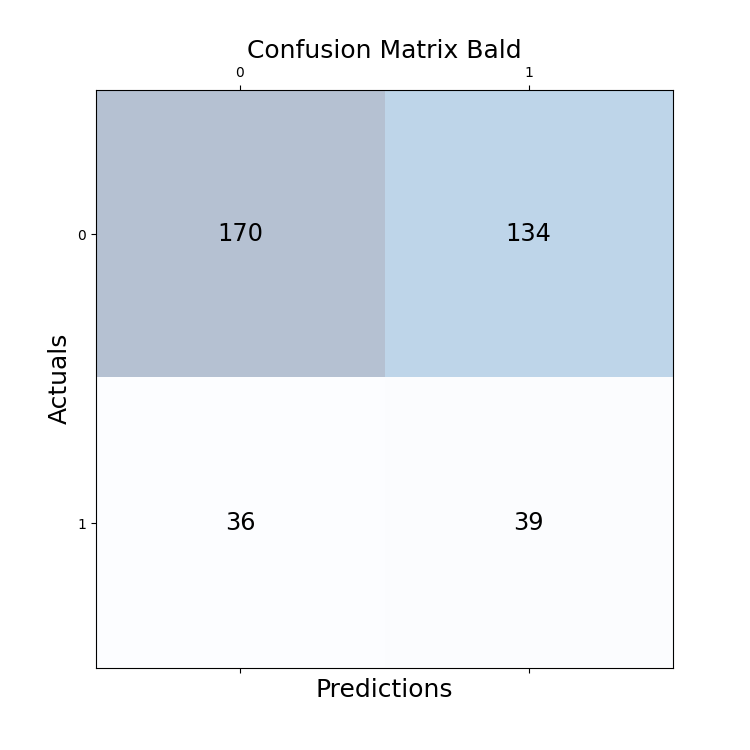}
    \caption{Confusion Matrix of the Performance of best BALD model}
    \label{baldcm}
\end{figure}

\begin{figure}[!ht]
\includegraphics[width=0.5\textwidth]{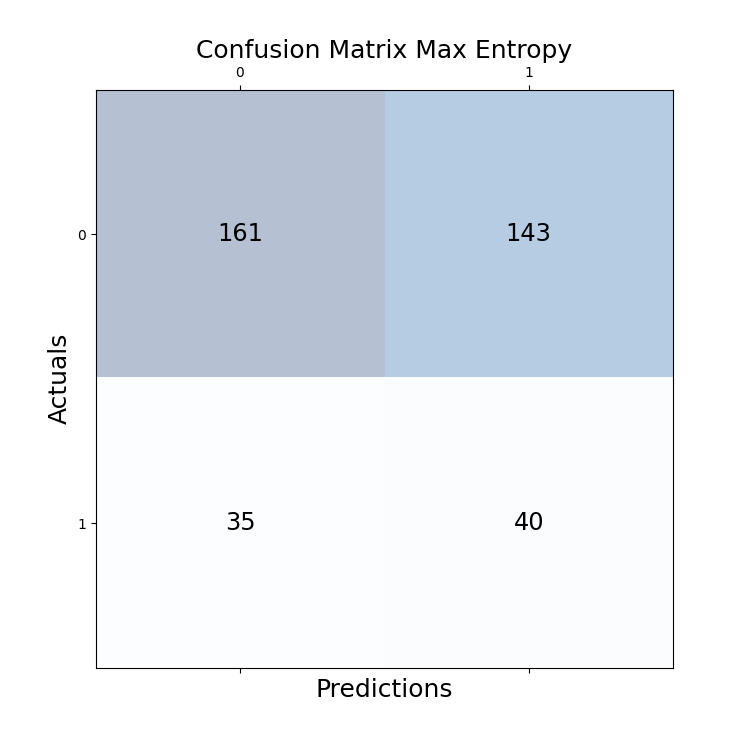}
    \caption{Confusion Matrix of the Performance of best \textit{max\_entropy} model}
    \label{maxentropy}
\end{figure}

\begin{figure}[!ht]
\includegraphics[width=0.5\textwidth]{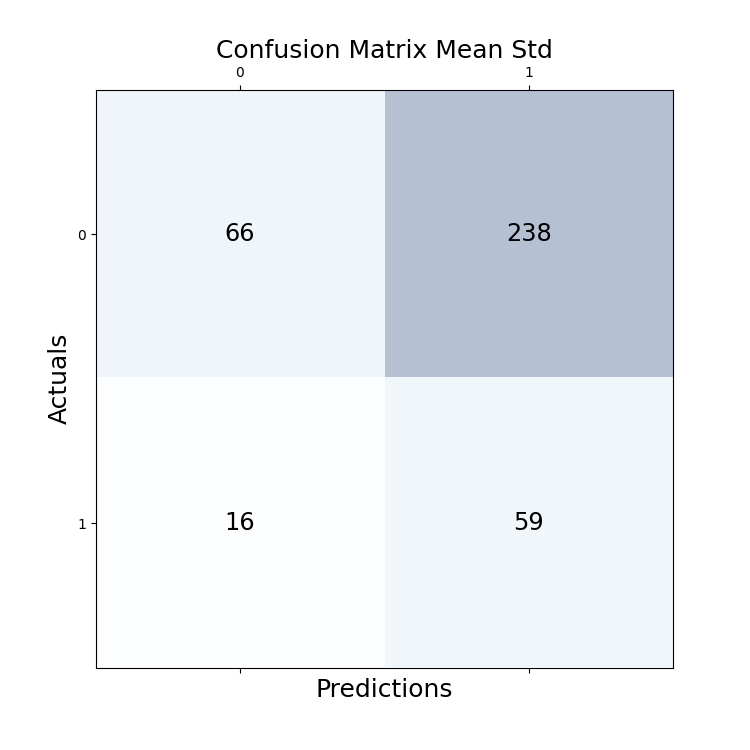}
    \caption{Confusion Matrix of the Performance of best \textit{mean\_std} model}
    \label{meanstd}
\end{figure}

We can observe a similar trend in Table \ref{query_size_effect}. We can see that the loss values are almost similar, while most of the acquisition functions achieved their highest accuracy scores around the original query size of 100 (except for mean\_std which performed better in terms of accuracy, with the second-lowest query size). This high classification accuracy despite the small and constrained training set, might be due to the capacity of the model, being able in $\sim$80\% of cases, to accurately distinguish between cancerous and non-cancerous images. However, if we look at the distribution of examples in each class, this level of accuracy could also be achieved if the model is correctly classified non-cancerous while misclassifying \textbf{all} cancerous samples. In order to verify our hypotheses, we proceeded and plotted three confusion matrices maps:
\begin{itemize}
    \item for \textit{bald}, we considered the model obtained with a query size of 100,
    \item for \textit{max\_entropy}, we considered the best as the one obtained with a query size of 70, offering the lowest testing loss, and a very competitive testing accuracy,
    \item for \textit{mean\_std}, we considered the best as the one obtained with a query size of 100, offering the lowest testing loss, and a very competitive testing accuracy.
\end{itemize}
The results are shown in Figure \ref{baldcm}, \ref{maxentropy}, and \ref{meanstd}.

The results show that \textit{bald} is more accurate at detecting non-cancerous samples. \textit{mean\_std} overfits a lot on cancerous samples, as our analyses suggested earlier, and classifies non-cancerous samples as cancerous. Maximum Entropy does also a decent job for the classification of both classes. These results on another hand could suggest that these methods behave a bit naively, exploiting the unbalance in the data. It is therefore paramount to explore new acquisition functions that could better behave with class unbalance while improving the performance on the downstream task. 
\section{Conclusion and Future Works}
In this work, we demonstrated how active learning could be used for a classification downstream task on the Melanoma Dataset. First of all, we showed that using uncertainty (epistemic) is useful for the Melanoma detection task. Next, we demonstrated that it is better for the model to query the most uncertain samples using the designated acquisition functions. Once that was settled, we leveraged several acquisition functions and found out that on average \textit{bald} performs the best, as authors \cite{bald} have claimed. However, our extended analyses have revealed that those acquisition functions behave naively, exploiting the data unbalance which in this case, would have had a huge impact on the classification accuracy if the majority class was the cancerous one. However, this also demonstrates, despite all the advantages and shortcomings of the different acquisitions we leveraged, that it is hard to work and generalize in an extremely low data regime. As future work, we could leverage how well these acquisition functions perform on later versions (and bigger) of the ISIC Dataset. Another interesting future is extending our work to the new acquisition function \textit{EPIG} introduced in \cite{smith2023predictionoriented}. EPIG measures information gain in the space of predictions rather than parameters and leads to a better performance than BALD.

{\small
\bibliographystyle{ieee_fullname}
\bibliography{egbib}
}
\end{document}